\documentclass[runningheads]{llncs}

\usepackage[T1]{fontenc}
\usepackage{graphicx}
\usepackage[table]{xcolor}
\usepackage[hidelinks]{hyperref}
\usepackage{multirow}
\usepackage{multicol}
\usepackage{booktabs}
\usepackage{rotating}
\usepackage{orcidlink}
\usepackage{subcaption}

\usepackage{tabularx}
\newcolumntype{Y}{>{\centering\arraybackslash}X}

\usepackage{color}

\usepackage{pifont}
\newcommand{\cmark}{\ding{51}}
\newcommand{\xmark}{\ding{55}}
\newcommand{\rmark}{\ding{81}}

\usepackage[disable,colorinlistoftodos,prependcaption,textsize=tiny]{todonotes}

\definecolor{light-gray}{gray}{0.95}
\definecolor{success}{RGB}{0, 128, 0}
\definecolor{failure}{RGB}{255, 0, 0}

\begin{document}
\title{SoccerDiffusion: Toward Learning End-to-End Humanoid Robot Soccer from Gameplay Recordings}
\titlerunning{SoccerDiffusion: End-to-End Humanoid Robot Soccer}
\author{
    Florian Vahl$^*$\orcidlink{0009-0001-5169-5864} \and
    Jörn Griepenburg$^*$\orcidlink{0009-0001-1602-1653} \and
    Jan Gutsche$^*$\orcidlink{0000-0002-0868-8589} \and \\
    Jasper Güldenstein\orcidlink{0000-0002-5416-5850} \and
    Jianwei Zhang\orcidlink{0000-0002-7856-5760}
}

\authorrunning{F. Vahl et al.}

\institute{University of Hamburg, Vogt-Kölln-Straße 30, 22527 Hamburg, Germany\\
    \email{\{florian.vahl, joern.griepenburg, jan.gutsche\}@studium.uni-hamburg.de, \{jasper.gueldenstein, jianwei.zhang\}@uni-hamburg.de}
}

\maketitle
\def\thefootnote{*}\footnotetext{These authors contributed equally to this work.}\def\thefootnote{\arabic{footnote}}

\begin{abstract}
\label{sec:abstract}

This paper introduces SoccerDiffusion, a transformer-based diffusion model designed to learn end-to-end control policies for humanoid robot soccer directly from real-world gameplay recordings.
Using data collected from RoboCup competitions, the model predicts joint command trajectories from multi-modal sensor inputs, including vision, proprioception, and game state.
We employ a distillation technique to enable real-time inference on embedded platforms that reduces the multistep diffusion process to a single step.
Our results demonstrate the model’s ability to replicate complex motion behaviors such as walking, kicking, and fall recovery both in simulation and on physical robots.
Although high-level tactical behavior remains limited, this work provides a robust foundation for subsequent reinforcement learning or preference optimization methods.
We release the dataset, pretrained models, and code under: \url{https://bit-bots.github.io/SoccerDiffusion}

\end{abstract}

\section{Introduction}
\label{sec:introduction}

Developing autonomous robots capable of effective and adaptive behavior, particularly in highly dynamic environments, has been a significant area of research.
One prominent example is RoboCup's Humanoid Kid-size League, where teams of four robots compete in fully autonomous soccer matches~\cite{rossiHumanLoopPerspectives2024}.
Manually engineering abstractions and interfaces between many components of a robotic system introduces a strong inductive bias, but it also limits the system's expressiveness.
Additionally, designing and integrating the subsystems into a coherent whole is a complex task.
The system complexity increases further when many edge cases and dynamics must be considered.
Therefore, learning an end-to-end control system for autonomous robots is a promising but still challenging approach.
Tight integration of dynamic motion skills and high-level decision-making in an end-to-end way has been shown to be possible in simulation~\cite{deepmind_sim} and in simplified real-world robot soccer~\cite{deepmind_real,tirumala2024learning}.
Although these are end-to-end approaches, they often require a certain number of demonstrations for basic skills to bootstrap the initial learning process.
These demonstrations are typically handcrafted and tailored to the specific task at hand.

In contrast, we are interested in learning from a large amount of data that was collected during real-world competitions.
We use this data to train a base model that mimics the typical behavior of the robots in the RoboCup Humanoid League.
The goal is not to outperform the current software stack with this base model, but to investigate whether useful behaviors can be learned from the existing dataset.
Thereby, we can leverage the existing data to bootstrap further learning processes, such as reinforcement learning (RL) or preference optimization (PO), by providing a base model.

To train a transformer-based diffusion model, we utilize existing data in the form of on-robot recordings from the Hamburg Bit-Bots RoboCup team at official matches.
Sampling this model allows us to generate various behaviors that are consistent with the recorded data.
The model works in an end-to-end manner, meaning it takes sensor data as input and generates joint commands as output.
To improve the performance of the model, we use a distillation technique to reduce the multistep diffusion process to a single inference step.

\section{Related Work}
\label{sec:related_work}

\textbf{Imitation Learning} (IM), also known as "\emph{Learning from Demonstration}" (LfD), is an approach where an agent learns expert demonstrations without relying on rewards from the environment.
In the domain of control in highly dynamic environments, such as humanoid robotics, pure optimization-based approaches often face challenges.
The high dimensionality of state and action spaces and complexity in dynamics make it highly difficult to design effective reward functions~\cite{imitation_learning_survey_humanoid}.
It has been shown that having prior understanding in the form of demonstrations is more effective than pure reinforcement learning~\cite{learning_from_demonstration,imitation_learning_survey}.
For these reasons, it has long been considered a promising approach in the field of humanoid robotics.
\\\\
\textbf{Behavioral Cloning} (BC) is a simple form of imitation learning, where the agent learns a policy by mapping state-action pairs based on the given demonstrations \cite{alvinn_behavior_cloning,framework_for_behavior_cloning}.
Because of its simplicity and effectiveness, BC has been widely used in various domains (robotics, autonomous driving, and game playing~\cite{imitation_learning_survey}).
The training data of states (\emph{input}) and actions (\emph{output}) is used to extract and replicate the demonstrator's policy by classification or regression, depending on the nature of the action space (\emph{discrete} vs. \emph{continuous}).
However, using simple modeling choices for BC has been shown to have limitations.

As Ross and Bagnell~\cite{reduction_for_imitation_learning} have shown, this classical supervised learning method can fall short for sequence prediction tasks.
Since learned predictions affect future states during policy execution, this violates the i.i.d. assumptions of most statistical learning methods.
Which leads to a compounding error and a \emph{distributional shift} problem, where the agent might result in a state that was not covered by the demonstration data.

Additionally, in situations where there is multimodality of the action space at decision points (multiple possible continuous actions for a given state), different modeling choices can lead to issues \cite{pearce2023imitating}.
Using simple point estimates (e.g., \emph{MSE}) can lead to a loss of diversity and 'average' policies, which produce actions that, while being close to the mean, are unlikely in the original distribution.
\emph{Discretising} the action space can introduce \emph{quantization errors}, lose dependencies between dimensions, and lead to 'uncoordinated' behavior.
As shown, for these purposes, diffusion models are a promising alternative to classical BC methods.
\\\\
\textbf{Transformer Models} have recently gained significant attention in the field of deep learning, particularly in natural language processing (NLP) and computer vision~\cite{transformers_attention_is_all_you_need}.
More recently, they have also emerged in the field of robotics~\cite{wolf2025diffusionmodelsroboticmanipulation}.
They have been shown to be highly effective in capturing long-range dependencies and complex relationships within data.
These capabilities make them suitable for robotic tasks, which often involve sequential data over long time horizons.
\\\\
\textbf{Diffusion Models} (DM) have been utilized for various generative and manipulative multi-modal tasks (e.g., text-to-image, -video, and -audio or image-inpainting) in recent years~\cite{diffusion_model_design_survey}.
DMs are a class of generative models that learn to generate data by reversing a diffusion process, which gradually adds noise to the data until it becomes indistinguishable from random noise \cite{DDPM}.
By learning to reverse this process, diffusion models can generate plausible samples from complex distributions.
Besides the above modalities, DMs have been successfully applied to robotics as a novel IM approach.
These work by imitating actions from expert demonstrations given a diffused state~\cite{reuss2023goal,pearce2023imitating,chi2024diffusionpolicy}.
Without downplaying the successes of these works, their application and model evaluation is limited to short multistep tasks, mostly for robot arms in simulated or real-world environments.
In contrast, we aim to apply DMs to a more complex and long-term task, such as humanoid robot soccer.
Similarly to Chen et al.~\cite{diffusion_behavior_cloning}, we do not intend to use a diffusion model as our principal policy, but as an intermediary model for other BC methods.

\section{Approach}
\label{sec:approach}

Our approach involves leveraging a transformer-based diffusion model to learn complex behaviors from recorded gameplay data.
By using a data-driven methodology, we aim to determine if a game play behavior can emerge from the dataset without explicit rule-based programming.
The motivation for using a diffusion model stems from its capacity to capture multimodal distributions, making it particularly well-suited for scenarios where multiple plausible outcomes exist.

The primary objective of this study is to investigate whether useful behaviors can be learned from the existing dataset rather than surpassing the performance of the current, manually programmed software stack.
This study is primarily academic in nature, focusing on the feasibility of behavior learning from available data rather than serving as a direct replacement for existing control algorithms.
In addition to this, we also aim to provide pre-trained models and a dataset to the community, serving as a foundation for future  studies (e.g., RL).

The training pipeline begins with the collection and preprocessing of data, followed by the training of a transformer-based neural architecture capable of learning behavior representations.
Finally, to facilitate real-time inference on resource-constrained hardware like many humanoid platforms, we employ a distillation technique that reduces the multistep diffusion process to a single inference step.

\subsection{Data Collection}
\label{sec:data_collection}

We utilize data from the Hamburg Bit-Bots team that was collected using their conventional software stack during the competitions RoboCup 2024\footnote{\url{https://data.bit-bots.de/ROSbags/robocup_2024/}} as well as RoboCup German Open 2025\footnote{\url{https://data.bit-bots.de/ROSbags/german_open_2025/}}.
In total, we considered 88 recordings, amounting to roughly 15 hours of data.
Other recordings from that team were dismissed because the image data was only stored once per second.

We define a recording as a single sequence of data collected from a single robot during a match.
During a match, the robot's software may be restarted multiple times, which results in multiple recordings.
A match may contain multiple recordings from multiple robots.

The raw data is stored as a \textit{ROS 2 Bag} in the \textit{mcap}~\cite{MCAP} file format and contains sensor measurements, intermediary representations, debug information, and finally the joint commands sent to the servo motors.
We reached out to other RoboCup soccer teams and got additional raw data from their robots.
However, due to time constraints, we were unable to include this data in the evaluation for this paper.

\subsection{Data Preprocessing}
\label{sec:data_preprocessing}

We preprocessed the data from the \textit{mcap} files to ensure that all modalities were resampled at the same rate and in a synchronized manner.
This simplified later processing, reduced the data volume, and enabled a uniform input format.

We resampled and synchronized all joints and IMU-related data at 50 Hz (the original data was sampled at over 300 Hz).
A sampling rate of 10 Hz was used for the images.
To avoid any causality issues and data leakage from future states during resampling, no interpolation with future data was performed.
Instead, we opted for selecting the last available data point before the sampling time.
The images were downsampled to $480\times480$ pixels.
The game state messages, as received from the human-controlled game controller software, were simplified to three states representing whether the robot is allowed to play (and move) or not, or unknown.
The resulting data was written to a \textit{SQLite} database,
which allowed us to query the data efficiently in a synchronized manner during training.

Due to an error at the RoboCup 2024 competition, the IMU data was not directly recorded.
Instead, the IMU data was reconstructed from an intermediary representation.
The reconstruction was successful, but it only allowed us to extract the robot's roll and pitch orientation.
Data for the accelerometer and gyroscope are unavailable and, therefore, not used in the model.

In \autoref{tab:database_statistics}, we can see the kind of data that was preprocessed, sampled, and finally stored in a single database file.
Additionally, we show some statistics about the dataset.
\vspace{-1em}

\renewcommand{\arraystretch}{1.1}
\setlength{\tabcolsep}{1ex}
\setlength{\aboverulesep}{0pt}
\setlength{\belowrulesep}{0pt}
\begin{table}[ht]
    \centering
    \caption{
        Dataset Statistics: The table shows the duration and sample counts of the dataset used for training the model.
        In total, 88 recordings were used. The dataset amounts to 340 GB in size.
        \vspace{0.5em}
    }
    \label{tab:database_statistics}
    \begin{tabularx}{0.94\textwidth}{lcccc}
        \toprule
        \textbf{Metric} & \textbf{Minimum} & \textbf{Mean (±SD)} & \textbf{Maximum} & \textbf{Total}\\
        \midrule
        Duration [min] & 0.59 & 10.32±11.10 & 76.65 & 908.23\\
        \midrule
        Image samples & 200 & 5952±6638 & 45954 & 523819\\
        Rotation samples & 1014 & 30147±33378 & 229777 & 2652975\\
        Joint state samples & 1014 & 30147±33378 & 229777 & 2652975\\
        Joint command samples & 1014 & 30147±33378 & 229777 & 2652975\\
        Game state samples & 1 & 939±1001 & 6280 & 82638\\
        \bottomrule
        Total number of samples & 3273 & 97334±107712 & 741565 & 8565382\\
        \bottomrule
    \end{tabularx}
\end{table}
\vspace{-2em}

\subsection{Model Architecture}
\label{sec:model_architecture}

To learn useful behaviors from our data, we utilize a transformer neural network trained with a diffusion denoising objective.
We choose the diffusion model as it has been shown to be able to learn complex distributions.
Our data includes some ambiguous situations that are difficult to model with a simple predictive architecture,
as such models are more prone to collapsing onto the mean of the data distribution~\cite{pearce2023imitating,chi2024diffusionpolicy}.
This inherently makes the problem multimodal, which is a good fit for the diffusion model.
The transformer architecture is chosen due to its ability to model temporal dependencies,
combine multiple modalities, and its success in various domains, including robotics~\cite{wolf2025diffusionmodelsroboticmanipulation}.
Following the results from Pearce et al. ~\cite{pearce2023imitating}, we opt out of using classifier-free guidance~\cite{ho2022classifier}, which is often used in diffusion models.

\begin{figure}
    \centering
    \includegraphics[width=0.95\textwidth]{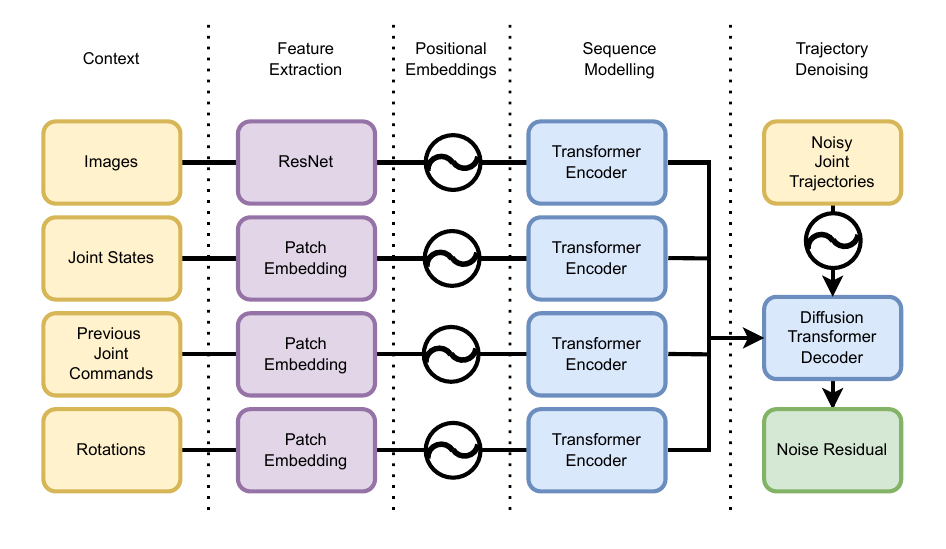}
    \caption{Architecture of the SoccerDiffusion model. Special tokens for the game state and the diffusion step are omitted for clarity.}
    \label{fig:model_architecture}
\end{figure}

Our model uses multiple encoders and a single joint command trajectory decoder, as shown in \autoref{fig:model_architecture}.
The encoders encode the joint states, previous joint commands, rotations, and images into a shared latent space.
They, therefore, perform a late fusion of the different modalities.
Their context window includes 2 seconds for the joint data and rotations and 1 second for the images.
The decoder predicts the noise residual given the latent encoder information and the noisy future joint command trajectory.

The joint and rotation encoders perform a linear token embedding of input patches, followed by a transformer encoder~\cite{transformers_attention_is_all_you_need}.
The image encoder uses a ResNet~\cite{resnet} architecture to extract features from the images before applying a transformer encoder to model the temporal dependencies.
In addition to the sensor data, we include both the step of the diffusion process as a sinusoidal embedding and the current game state as a learned token.

The joint command trajectory decoder uses a transformer decoder with cross-attention to the shared latent space of the encoders.
The context information is embedded once into the latent space, and the decoder queries the latent space for each denoising step.
The denoising diffusion implicit models (DDIM)~\cite{ddim} sampling algorithm is used to sample the joint command trajectory.
All transformers utilized sinusoidal position encodings to add temporal information to the tokens~\cite{transformers_attention_is_all_you_need}.

The model size is easily scalable; we selected hyperparameters (\autoref{tab:hyperparams}),
to enable single-step inference on the Ryzen 7 5700U APU used in the \textit{Wolfgang-OP}~\cite{wolfgang} robot,
leveraging the distillation process from \autoref{sec:distillation}.
We use the same model architecture for the distillation process, only reducing the number of diffusion steps to 1.

\renewcommand{\arraystretch}{1.1}
\setlength{\tabcolsep}{1ex}
\setlength{\aboverulesep}{0pt}
\setlength{\belowrulesep}{0pt}
\begin{table}[ht]
    \centering
    \caption{
        Evaluation Hyperparameters (J = Joint, R = Rotation, L = Layers)
        \vspace{0.5em}
    }
    \label{tab:hyperparams}
    \begin{tabularx}{0.97\textwidth}{lr || lr}
        \toprule
        \textbf{Hyperparameters} & \textbf{Value} & \textbf{Hyperparameters} & \textbf{Value}  \\
        \midrule
        J. Transformer Encoder & 4L & Epochs & 300 \\
        R. Transformer Encoder & 4L & Rotation Representation & Axis angle~\cite{curtright2014compact} \\
        Image Transformer Encoder & 2L & Image Encoder & ResNet-18~\cite{resnet} \\
        Decoder Transformer & 4L & Image Size & 224×224 \\
        J. Transformer & 4L & J \& R Token Patch Size & 5 \\
        R. Transformer & 2L & Learning Rate & 1e-4 \\
        Decoder Transformer & 6L & Optimizer & AdamW~\cite{adamw} \\
        Image Transformer & 1L & Learning Rate Scheduler & Cosine \\
        Attention Heads & 8 & Diffusion Sampler & DDIM~\cite{ddim} \\
        Hidden Dimensions & 512 & Training Diffusion Steps & 1000 \\
        Batch Size & 256 & Sampling Diffusion Steps & 30 \\
        \bottomrule
    \end{tabularx}
\end{table}
\vspace{-1em}

\subsection{Distillation}
\label{sec:distillation}

To enable real-time inference while still matching a multimodal distribution, the multistep diffusion was distilled into a single-step diffusion model.
This allows the model to directly predict the joint command trajectory based on the latent encoder information and the input noise.
As pioneered by Luhmann et al.~\cite{luhman2021knowledge}, we use a teacher-student approach to distill the diffusion model.
The architecture of the student model is identical to the teacher model, but the number of diffusion steps is reduced to 1.
This allows us to initialize the student model with the weights of the teacher model.
During training, a deterministic 30-step DDIM sampling is performed to generate the target trajectory using the teacher model.
The student model is trained to predict the target trajectory given the input noise and the latent encoder information using a mean squared error loss.
During distillation, only the decoder is trained, while the encoder weights are frozen.
Both the teacher and student models utilize the same encoder latents to speed up the distillation process.
The distillation process is performed using the same dataset that was used to train the teacher model.

\subsection{Trajectory Execution}
\label{sec:trajectory_execution}
The generated joint command trajectory is executed step-wise in a delay compensating manner.
Similarly to the training process, the model's generated trajectory starts right after the latest available input data.
However, we must consider the model's inference time.
Once the model has finished the inference, the intended execution time of the first points of the generated trajectory has already passed.
We, therefore, only partially execute a generated trajectory, cutting off points from the beginning depending on the model runtime.
Trajectories are generated in an overlapping manner; we always execute from the most recent trajectory and send the joint command to the actuators.
This allows us to hold up the assumption of an instantaneous inference during training, without incorporating any execution delays into the model.

\section{Evaluation}
\label{sec:evaluation}

Due to the end-to-end nature of the model, it is difficult to quantify specific subsystems like the walking or kicking behavior.
Most of these subsystems are, therefore, evaluated qualitatively.
The evaluation is performed on the physical robot and in simulation using Webots~\cite{webots}. \todo{hier könnte man noch hlvs citen und sagen man benutzt das gleiche environment}
Evaluating the model in simulation allows us to quickly test the model and
also see how it performs when the domain is shifted.
A domain shift is present because the simulation is not part of the training data.
Still, the model is able to overcome this sim to real gap and perform similarly in simulation.

\subsection{Quantitative Evaluation}
\label{sec:quantitative_evaluation}

To quantify the performance of our approach, we performed a series of ten fall-recovery tests for all four distinct directions.
The robot was placed in a standing position and manually pushed to the ground in the desired direction.
The stand-up motion was performed by the SoccerDiffusion policy with an overall high success rate ($95\%$ physically, $100\%$ in simulation), and fall detection was successful in all cases.
This is compared to the baseline software stack with $100\%$ success rate (physically and in simulation).
We present the detailed results in \autoref{tab:stand_up_motion}.
\vspace{-1em}

\renewcommand{\arraystretch}{1.2}
\setlength{\tabcolsep}{1ex}
\setlength{\aboverulesep}{0pt}
\setlength{\belowrulesep}{0pt}
\begin{table}[h]
    \caption{
        Stand-up motion evaluation in the Webots simulator or on the physical robot, either executed with baseline code or SoccerDiffusion.
        We measured success if the robot was able to recover from a fall by itself on the first try~(\cmark) or if it took a few retries~(\rmark).
        A failure~(\xmark) happened when the robot was unable to recover from a fall by itself and needed to be manually reset.
        \vspace{0.5em}
    }
    \label{tab:stand_up_motion}
    \centering
    \begin{tabularx}{0.95\textwidth}{c || Y Y || Y Y | Y Y Y}
        \toprule
        \multicolumn{1}{c||}{\textbf{Fall direction}}
        & \multicolumn{2}{c||}{\textbf{Baseline}}
        & \multicolumn{2}{c|}{\textbf{Simulation}}
        & \multicolumn{3}{c}{\textbf{Physical}} \\
                              & {\cmark}
                              & {\xmark}
                              & {\cmark}
                              & {\xmark}
                              & {\cmark}
                              & {\rmark}
                              & {\xmark} \\
        \midrule
                              \textit{Forward}
                              & \textcolor{success}{10}
                              & -
                              & \textcolor{success}{10}
                              & -
                              & 6
                              & 3
                              & \textcolor{failure}{1} \\

                              \textit{Backward}
                              & \textcolor{success}{10}
                              & -
                              & \textcolor{success}{10}
                              & -
                              & \textcolor{success}{10}
                              & -
                              & - \\

                              \textit{Left side}
                              & \textcolor{success}{10}
                              & -
                              & \textcolor{success}{10}
                              & -
                              & \textcolor{success}{10}
                              & -
                              & - \\

                              \textit{Right side}
                              & \textcolor{success}{10}
                              & -
                              & \textcolor{success}{10}
                              & -
                              & 9
                              & -
                              & \textcolor{failure}{1} \\
        \bottomrule
    \end{tabularx}
\end{table}
\vspace{-2em}

\subsection{Qualitative Evaluation}
\label{sec:qualitative_evaluation}

For the qualitative evaluation, we used one \textit{Wolfgang-OP} robot in the real world.
The robot was able to walk in a stable manner and perform various motions, such as kicking, turning, standing up, and walking in all directions.
It showed dynamic motions and stabilization behaviors mimicking the way the Hamburg Bit-Bots stabilize their robot (PID-based Center of Mass control and adaptive gait phase) and is able to recover from disturbances.
However, it is difficult to quantify the performance for these motions due to the end-to-end nature of the model:
Evaluating the walking speed or the traversed distance is not meaningful because we cannot influence the model, e.g., to walk at maximum speed, and we do not know the intended behavior.

The robot's motions are very similar to the ones in the dataset, but the robot fails to show any high-level behavior, such as intentional kicking or positioning itself on the soccer field.
Possible reasons could be a limited number of demonstrations, the model size, or the data distribution itself.
Due to team play, not all players are moving towards the ball at all times, which increases the complexity of the dataset and might be difficult to learn for the model.

\begin{figure}[ht]
    \centering
    \begin{subfigure}{0.99\textwidth}
        \centering
        \includegraphics[width=\textwidth]{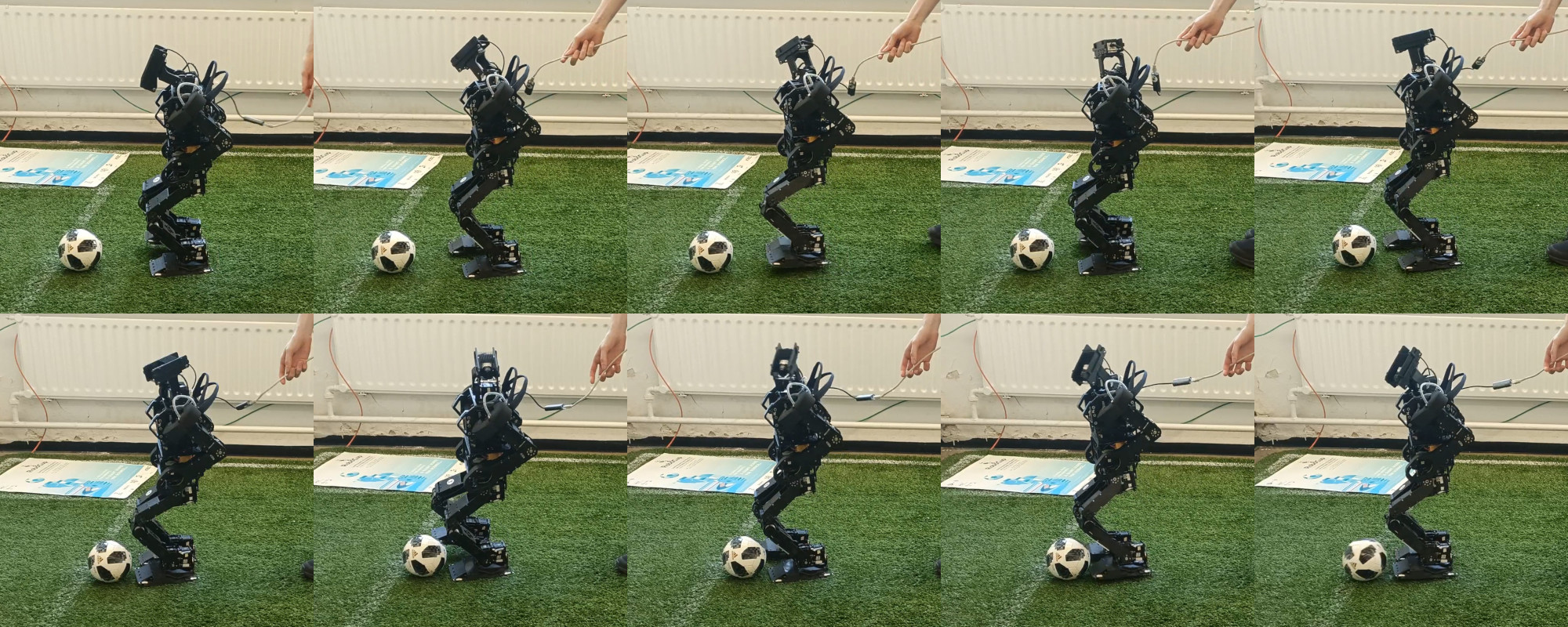}
        \caption{Bipedal walk cycle}
        \label{fig:qualitative_evaluation_walk}
    \end{subfigure}\\[1ex]
    \begin{subfigure}{0.99\textwidth}
        \centering
        \includegraphics[width=\textwidth]{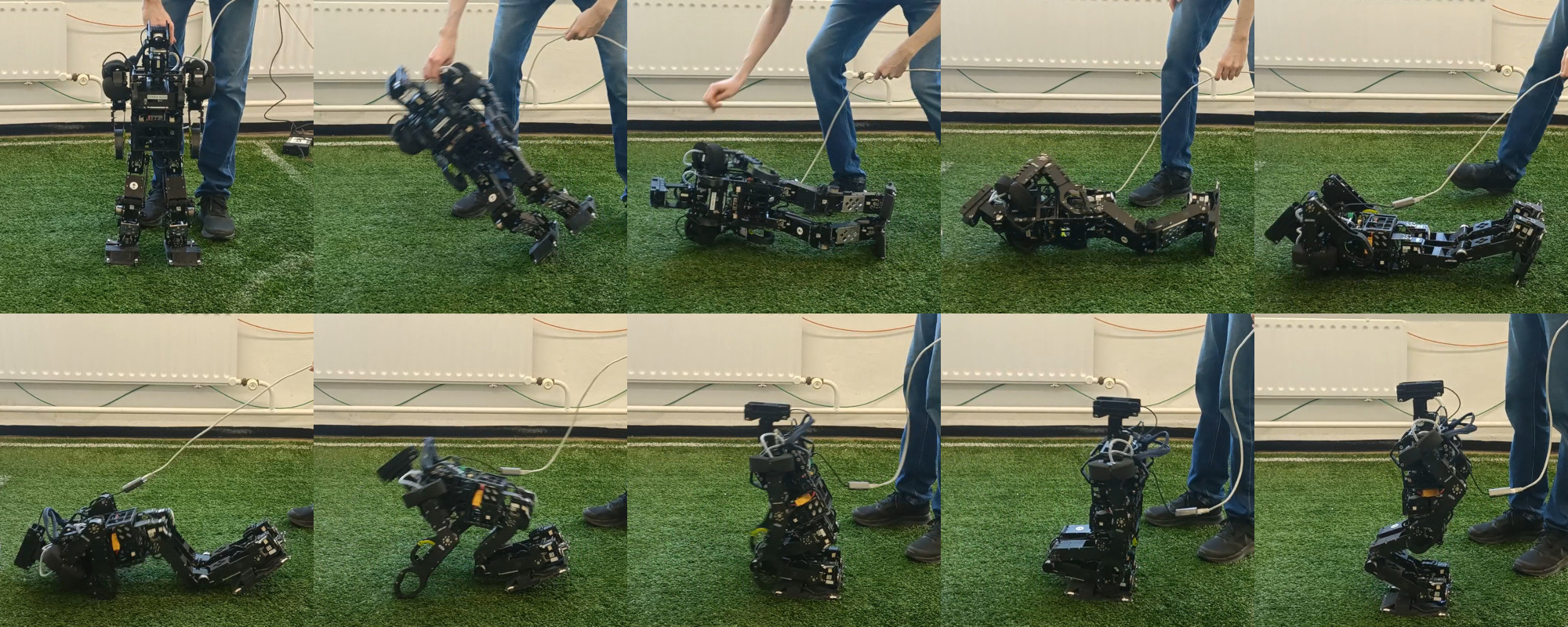}
        \caption{Fall recovery from the right side}
        \label{fig:qualitative_evaluation_fall}
    \end{subfigure}
    \caption{Qualitative evaluation: (a) walking and (b) fall recovery, both  performed by the SoccerDiffusion policy on the \textit{Wolfgang-OP} platform \cite{wolfgang}.}
    \label{fig:qualitative_evaluation_grouped}
\end{figure}

The image sequence in \autoref{fig:qualitative_evaluation_grouped} shows the robot walking forward and recovering from a fall to the side, both using our SoccerDiffusion policy.
When executing the falling and standing up motion, the robot performs a multistep motion (i.e., turning itself from the side to the front and then standing up).

We did not observe a qualitative difference between the distilled and the non-distilled model running in a sub-real-time simulation environment.

Different head motions imitating the training data can be observed.
Firstly, the robot performs a head-scan motion to the left and right side at various tilt angles.
This is usually done to scan the environment for more accurate localization.
Secondly, the robot tilts the head to observe the ball and the environment.
This behavior is mimicking the head behavior during ball manipulation in the dataset.
The robot does not follow the ball with its head, which is to be expected since robots in the training data do not track the ball either.

A major limitation of the current model is a strong bias to keep standing still, if there was no prior motion in the context window.
This is likely because the dataset contains many sequences of standing robots.
Together with a lack of high-level behavior, this leads to the robot rarely starting to walk on its own.

Overall, the learned motions are seemingly very robust, as the model can overcome the gap between the real world and simulation without any additional training on simulated data (see \autoref{tab:stand_up_motion}).
The software stack of the Hamburg Bit-Bots interestingly does not exhibit this behavior, as the walk engine parameters need to be tuned for the simulation.

\section{Future Work}
\label{sec:discussion_future_work}

Since we could not reach the same level of abilities as the conventional baseline, we recognize significant potential for future improvements.
Currently, we are expanding our dataset with recordings from other teams and 3D-simulated autonomous matches.
Furthermore, employing a general-purpose vision embedding model such as DINOv2~\cite{oquab2023dinov2} could facilitate the learning of more complex, high-level behaviors while also reducing the computational demands of training, as the embeddings can be precomputed.
From experience, we know the dataset contains long sequences of a robot standing still during the game.
We could filter out these sequences to encourage more dynamic actions.
While this is not the primary focus of this work, auxiliary losses for predicting intermediate representations, such as the robot's position or walk velocity, could be beneficial.
A detailed ablation study could identify more effective configurations for our task.

\section{Conclusion}
\label{sec:conclusion}

This work explored the potential of learning end-to-end policies for the RoboCup Humanoid League context.
Compared to previous work, this approach learns directly from gameplay recordings, which are relatively easy to obtain.
Our approach leverages the strengths of diffusion models in multimodal output distributions with the strengths of transformers in modeling temporal dependencies.
We hypothesize that this combination of architectures is well-suited for the unstructured nature of the data.

Our model successfully learned to reproduce a range of fundamental behaviors from the dataset.
The generated motions included stable walking in various directions, turning, kicking motions, and robust fall recovery sequences.
Notably, the model replicates stabilization behaviors to maintain balance while walking.
Furthermore, we show real-to-sim transfer of the learned model.
We achieved real-time performance by lowering computational costs through distillation.

We laid a foundation for further optimization of an end-to-end policy for humanoid robot soccer using RL or PO techniques.
However, we did not achieve the same level of behavior as the Hamburg Bit-Bots, such as intentional kicking or positioning on the field.
We suspect this is due to the limited dataset or training duration and will be addressed in future work.

\begin{credits}
\subsubsection{\ackname}
We gratefully acknowledge funding and support from the project \textit{Digital and Data Literacy in Teaching Lab (DDLitLab)} at the University of Hamburg and the \textit{Stiftung Innovation in der Hochschullehre} foundation.
We extend our special thanks to the members of the \textit{Hamburg Bit-Bots} RoboCup team for their continuous support and for providing data and computational resources.
We also thank the RoboCup teams \textit{B-Human} and \textit{HULKs} for generously sharing their data for this research.
Additionally, we are grateful to the \textit{Technical Aspects of Multimodal Systems (TAMS)} research group at the University of Hamburg for providing computational resources.
This research was partially funded by the Ministry of Science, Research and Equalities of Hamburg, as well as the German Research Foundation and the National Science Foundation of China through the project \textit{Crossmodal Learning} (TRR-169).

\subsubsection{\discintname}
The authors have no competing interests to declare that are relevant to the content of this article.
\end{credits}

%
%
%
\bibliographystyle{splncs04}
\bibliography{references}

\end{document}